\documentclass[sigconf,nonacm]{acmart}

\AtBeginDocument{}
\setcopyright{none}
\acmDOI{}
\acmISBN{}
\acmYear{2025}
\copyrightyear{2025}
\acmConference[AI4DF'25 (Review)]{AI and Data Science for Digital Finance Workshop}{Nov 15, 2025}{Singapore}
\acmSubmissionID{AI4DF-XXXX}

\usepackage{graphicx}
\usepackage{amsmath,amssymb}
\usepackage{pifont}
\newcommand{\cmark}{\ding{51}}

\usepackage{multirow}
\usepackage{enumitem}
\usepackage{hyperref}
\usepackage{float}
\usepackage{url}
\newcommand{\rundir}{output/run_20250808_221301}
\newcommand{\robustdir}{output/robust_output/compare_20251001_013842}

\DeclareRobustCommand{\showtt}[1]{%
  \begingroup
    \edef\temp{#1}
    \expandafter\path\expandafter{\temp}
  \endgroup
}

\graphicspath{{\rundir/}{\robustdir/}{./}}
\DeclareGraphicsExtensions{.pdf,.png,.jpg}

\makeatletter
\newcommand{\safeinclude}[2][]{%
  \begingroup
    \def\A{\rundir/#2}\def\B{\robustdir/#2}\def\C{#2}%
    \IfFileExists{\A}{\includegraphics[#1]{\A}}{%
      \IfFileExists{\B}{\includegraphics[#1]{\B}}{%
        \IfFileExists{\C}{\includegraphics[#1]{\C}}{%
          \fbox{\parbox{.95\columnwidth}{\centering\small Missing figure: \texttt{\detokenize{#2}}}}}}}%
  \endgroup
}
\newcommand{\safeinput}[1]{%
  \IfFileExists{#1}{\input{#1}}{%
    \begin{table}[t]\centering
      \caption{(Auto placeholder) table not found at compile time.}
      \label{tab:auto-missing}
      \begin{tabular}{@{}lc@{}}\toprule Item & Note\\\midrule layerC & table file missing\\\bottomrule\end{tabular}
    \end{table}}
}
\makeatother

\title{Forecasting Crypto Market Risk via Liquidity Spillovers:\\
A Multi-Layer Granger--VAR--HAR-X--ML Pipeline}

\title{A Multi-Layer ML–Econometric Pipeline for Forecasting Market Risk:
Evidence from Crypto Liquidity Spillovers}

\author{Yimeng Qiu}
\affiliation{\institution{LSEG}\country{US}}

\author{Feihuang Fang}
\affiliation{\institution{LSEG}\country{UK}}

\begin{document}

\begin{abstract}
We test whether liquidity/volatility proxies of a small set of core cryptoassets exhibit spillovers that forecast \emph{market-wide} risk. Our pipeline stacks three statistical layers (A: core LV$\leftrightarrow$core returns; B: PC(LV)$\leftrightarrow$PC(returns); C: vol-PCs$\to$cross-sectional ``volatility crowding'') and complements them with VAR IRF/FEVD \cite{granger1969,sims1980}, HAR-X \cite{corsi2009simple}, and a leakage-safe ML protocol (time splits, early stopping, validation-only thresholding, SHAP).
Using daily data from 2021 to 2025 ($1462\times 74$; latest artifacts under \showtt{\rundir}), we find significant Granger links across layers and moderate OOS performance. We report only the most probative figures (pipeline overview, Layer~A heatmap, Layer~C robustness, VAR FEVD, and a test-set PR curve) while providing full CSV/PNGs in the artifact directory.
\end{abstract}

\maketitle

\section{Introduction}
Liquidity and volatility co-move across assets, producing spillovers during stress periods \cite{amihud2002,chordia2000commonality,diebold2012better,barunik2018measuring}. We ask whether a compact set of \emph{core} coins contains enough information to forecast \emph{market-wide} risk. We operationalize a three-layer chain of evidence:
\begin{itemize}[leftmargin=1.05em]
    \item \textbf{Layer A}: core liquidity/volatility (LV) $\leftrightarrow$ core returns.
    \item \textbf{Layer B}: principal components of LV $\leftrightarrow$ principal components of returns.
    \item \textbf{Layer C}: volatility PCs $\to$ a cross-sectional ``volatility crowding'' target.
\end{itemize}
We then estimate compact VARs for IRF/FEVD interpretation, fit HAR-X with HAC errors, and train early-stopped XGBoost models on chronological splits. Contributions: (i) a cross-sectional L2 ``crowding'' target aligned to vol-PCs; (ii) a leakage-aware forecasting protocol; (iii) reproducible artifacts (logs/CSVs/figures) under \showtt{\rundir}.

\section{Related Work}
We build on classical tools for dynamic dependence and volatility modeling. Granger causality and VARs provide the workhorse framework for testing predictive links and impulse propagation \cite{granger1969,sims1980}. Our liquidity/volatility proxies follow standard microstructure and range-based measures—Amihud illiquidity and Parkinson high/low volatility—and are complemented by GARCH-type variation \cite{amihud2002,parkinson1980,bollerslev1986}. To capture long-memory structure, we adopt HAR-X with HAC inference \cite{corsi2009simple,newey1987}. For forecasting and interpretability, we employ gradient-boosted trees and SHAP explanations \cite{chen2016xgboost,lundberg2017unified}. Relative to this literature, we connect \emph{core-coin} microstructure signals to a \emph{market-wide} cross-sectional ``volatility crowding'' target via a multi-layer causality chain, and we evaluate predictability under a leakage-aware protocol (see Figure 1).

\section{Method}
\subsection{Data and Preprocessing}
We assemble daily prices/volume from public aggregators (Yahoo Finance, CoinMarketCap) and build an engineered panel with $1462\times 74$ entries (latest run under \showtt{\rundir}). The core coin set is \textsc{ETH}, \textsc{BTC}, \textsc{YFI}, \textsc{DOT}, \textsc{XEM}, \textsc{BNB}, \textsc{ARK}. From these, we compute log-returns, Amihud illiquidity \cite{amihud2002}, turnover, and two volatility proxies (GARCH(1,1), Parkinson high/low). For interpretability, we retain key raw indicators for major cores and form principal components for the rest; the returns PCs keep $k{=}3$ explaining $85.26\%$ cumulative variance. To attenuate mechanical leverage effects, for each core we regress volatility proxies on its own return and keep residuals (residual vol/turnover columns). For comparison, we also construct legacy market targets based on $14$-day realized volatility and its rolling-standardized variant (\texttt{market\_rv14}, \texttt{market\_rv14\_rs}); all artifacts/CSVs are exported under \showtt{\rundir}.

\textbf{Core-coin selection.} We endogenously select ``core'' coins by constructing a directed Granger network on standardized log-returns and keeping the strongest \emph{senders} by PageRank \cite{brin1998pagerank}. Specifically,
\[
i \to j \ \text{iff}\ \min_{\ell=1,\dots,5} p_{i\to j}(\ell) < 0.05,
\]
and we rank nodes by PageRank on $G^\top$ (damping $\alpha{=}0.85$), taking the top-$N$ and enforcing a small whitelist $\{\textsc{BTC},\textsc{ETH},\textsc{BNB}\}$ for coverage.
The latest run yields \textsc{ETH}, \textsc{BTC}, \textsc{YFI}, \textsc{DOT}, \textsc{XEM}, \textsc{BNB}, \textsc{ARK}.
From these, we compute log-returns, Amihud illiquidity, turnover, and two volatility proxies (GARCH(1,1), Parkinson high/low). For interpretability, we retain key raw indicators for major cores and form principal components for the rest; the returns PCs keep $k{=}3$ explaining $85.26\%$ cumulative variance. To attenuate mechanical leverage effects, for each core we regress volatility proxies on its own return and keep residuals. We also construct legacy targets (\texttt{market\_rv14}, \texttt{market\_rv14\_rs}); all artifacts/CSVs are under \showtt{\rundir}.

\subsection{Targets and Factors}
From daily data on core coins (e.g., \textsc{BTC}, \textsc{ETH}, \textsc{BNB}, \textsc{YFI}, \dots), we compute log-returns, Amihud illiquidity, turnover, and high/low-based volatility. For each group, we form PCs. The \emph{volatility crowding} target is
\[
\textstyle \mathrm{mkt\_xsec\_vol\_l2}(t)=\sqrt{\frac{1}{K}\sum_{k=1}^{K} z_k(t)^2},\quad
z_k(t) = \frac{\mathrm{PC}^{(\mathrm{vol})}_k(t)-\mu_k}{\sigma_k},
\]
optionally rolling-standardized over a window $W$.

\begin{figure*}[t]
  \centering
  \safeinclude[width=\linewidth]{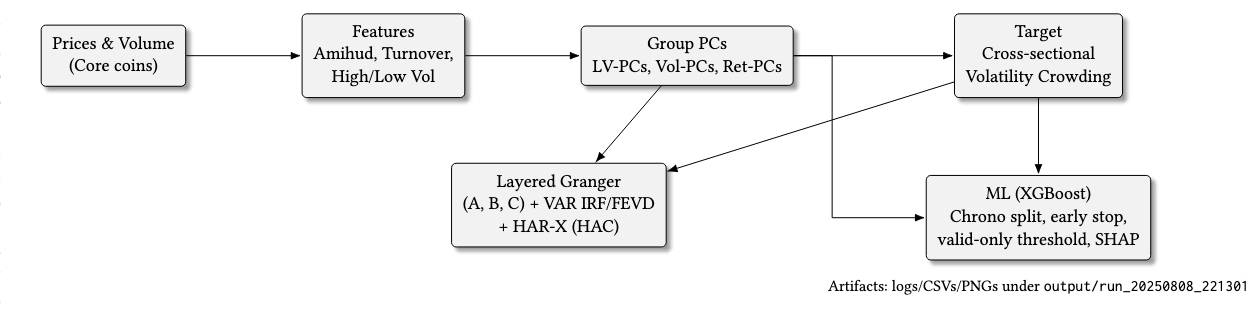}
  \caption{End-to-end pipeline: factor construction $\rightarrow$ layered causality (A/B/C) and interpretable VAR/HAR-X $\rightarrow$ leakage-safe ML evaluation.}
  \Description{Diagram showing the end-to-end pipeline from factor construction to layered causality analysis and ML evaluation.}
  \label{fig:pipeline}
\end{figure*}

\subsection{Layered Granger and VAR}
We run block-Granger tests \cite{granger1969} in small VARs \cite{sims1980} with BIC-selected order (fixed-lag variants for robustness). We report F, $p$, and BH-FDR $q$.
Two compact VARs are used for interpretation: (i) a ``main'' block (LV + two vol-PCs + target); (ii) a vol-only block (vol-PCs + target), with orthogonalized IRFs and FEVDs plus Ljung--Box $Q(10)$ checks. We implement block-Granger tests in small VARs (order by BIC, with fixed-lag variants for robustness) and report the F-statistic, $p$-value, and Benjamini--Hochberg FDR $q$ for each layer.

\subsection{HAR-X with HAC}
Following \cite{corsi2009simple}, we regress a market risk target $y_t$ (our cross-sectional crowding or RV variants) on daily/weekly/monthly averages of volatility PCs:
\begin{equation}
\label{eq:harx}
y_t=\beta_0+\beta_d\,\overline{v}_{t-1:t}+\beta_w\,\overline{v}_{t-5:t}+\beta_m\,\overline{v}_{t-22:t}+\varepsilon_t,
\end{equation}
where $\overline{v}_{t-a:t}$ denotes the mean of vol-PC regressors over the indicated window. Inference uses Newey–West HAC standard errors. 

\subsection{ML Protocol (Leakage-Safe)}
We construct an $H$-day risk index by blending short-horizon dispersion in features with structural shocks extracted from VARs. For $H{=}10$,
\begin{equation}
\label{eq:riskidx}
\mathrm{risk\_idx}_H(t)=\frac{1}{H}\sum_{h=1}^{H}\Big(\sigma^{(\mathrm{feat})}_{t-h+1}+\big|\epsilon^{(\mathrm{struct})}_{t-h+1}\big|\Big),
\end{equation}
where $\sigma^{(\mathrm{feat})}$ is a rolling standard deviation of selected inputs and $\epsilon^{(\mathrm{struct})}$ are rolling mean absolute structural shocks. We run chronological splits (70\%/15\%/15\% train/valid/test) with uniformly lagged features ($+1$ day) to ensure ``use today to forecast $t{+}H$''. Models are XGBoost \cite{chen2016xgboost} with early stopping; we tune only on validation. The classification threshold is \emph{chosen on the validation set} to maximize F1,
\begin{equation}
\hat{\tau}\in\arg\max_{\tau}\ \mathrm{F1}\big(\text{valid};\ \hat{p}(x)\ge\tau\big),
\end{equation}
then \emph{frozen} for the test set (positives correspond to the top 15\% risk regime). SHAP values \cite{lundberg2017unified} are computed on the fitted booster for attribution; full predictions/curves reside under \showtt{\rundir}.

\section{Experiments}\label{sec:experiments}
We use the latest run directory \showtt{\rundir} and robustness folder \showtt{\robustdir}.
Full tables/CSVs are exported there; we include only the most probative figures below \cite{liu2023forecasting,wang2023machine,qiu2021forecasting}.

\noindent\textbf{Key findings}
\emph{Layer A}: 25/35 block-Granger tests are significant at $p{<}.05$ with BH-FDR control; 
liquidity/volatility of \textsc{BNB}, \textsc{XEM}, and \textsc{YFI} lead their own returns, 
while returns Granger-cause residual vol/turnover for most cores.  

\emph{Layer B}: 5/15 PC-to-PC links are significant; examples include 
$\mathrm{pcLV}\!\to\!\mathrm{returns\_PC3}$ and 
$\mathrm{pcRET}\!\to\!\{\mathrm{amihud\_PC1},\mathrm{turnover\_PC1}, \\
\mathrm{garch\_vol\_PC3},\mathrm{park\_vol\_PC1}\}$.  

\emph{Layer C}: the raw market cross-sectional crowding target \showtt{mkt_xsec_vol_l2} 
is significantly predicted by volatility PCs at lag $3$ ($p{=}0.0146$); 
standardized or leave-$k$-out variants are not significant in the latest run.  

\emph{ML}: the risk-index regressor attains test $R^2{=}0.53$; 
the classifier reaches ROC-AUC $0.74$ and PR-AUC $0.47$, using a threshold tuned 
\emph{only} on validation and frozen on test.  

\emph{Baselines}: persistence-based regressions (AR/HAR-X) nearly saturate level prediction 
($R^2{\approx}0.998$), while in classification logistic regression achieves a higher ROC-AUC ($\sim$0.90) 
but lower PR-AUC ($\sim$0.44); our leakage-safe XGBoost instead delivers the best PR-AUC ($0.47$–$0.52$ across RS windows), 
which is the more informative measure under imbalance and demonstrates its superior early-warning ability.  
\begin{table}[H]
\centering
\caption{Task \& Metric \& Best Model (Test Value)}
\label{tab:baseline-best-vs-ours}
\begin{tabular}{@{}lll@{}}
\toprule
\midrule
\multirow{2}{*}{Regression} 
 & $R^2$ (test) & AR(H): $\sim 0.998$ \\
 & MSE (test)   & AR(H): $\sim 1.41$ \\
\midrule
\multirow{4}{*}{Classification} 
 & ROC--AUC (test) & Logit: $\sim 0.899$ \\
 & PR--AUC (test)  & \textbf{XGB: $\mathbf{0.522}$  (RS252)} \\
 & F1 (test)       & Logit: $\sim 0.52$ \\
 & Accuracy (test) & Logit: $\sim 0.90$ \\
\bottomrule
\end{tabular}
\end{table}

\clearpage
\noindent\textbf{Layer A and B.}
\begin{figure}[H]
  \centering
  \safeinclude[width=\columnwidth]{heatmap_layerA.png}
  \caption{Layer~A heatmap ($-\log_{10} p$). File: \showtt{\rundir/heatmap_layerA.png}.}
  \Description{Heatmap visualization for Layer A, showing $-\log_{10} p$ values.}
  \label{fig:heatA}
\end{figure}

\newpage
\noindent\textbf{Layer C robustness.}
\begin{figure}[H]
  \centering
  \safeinclude[width=\columnwidth]{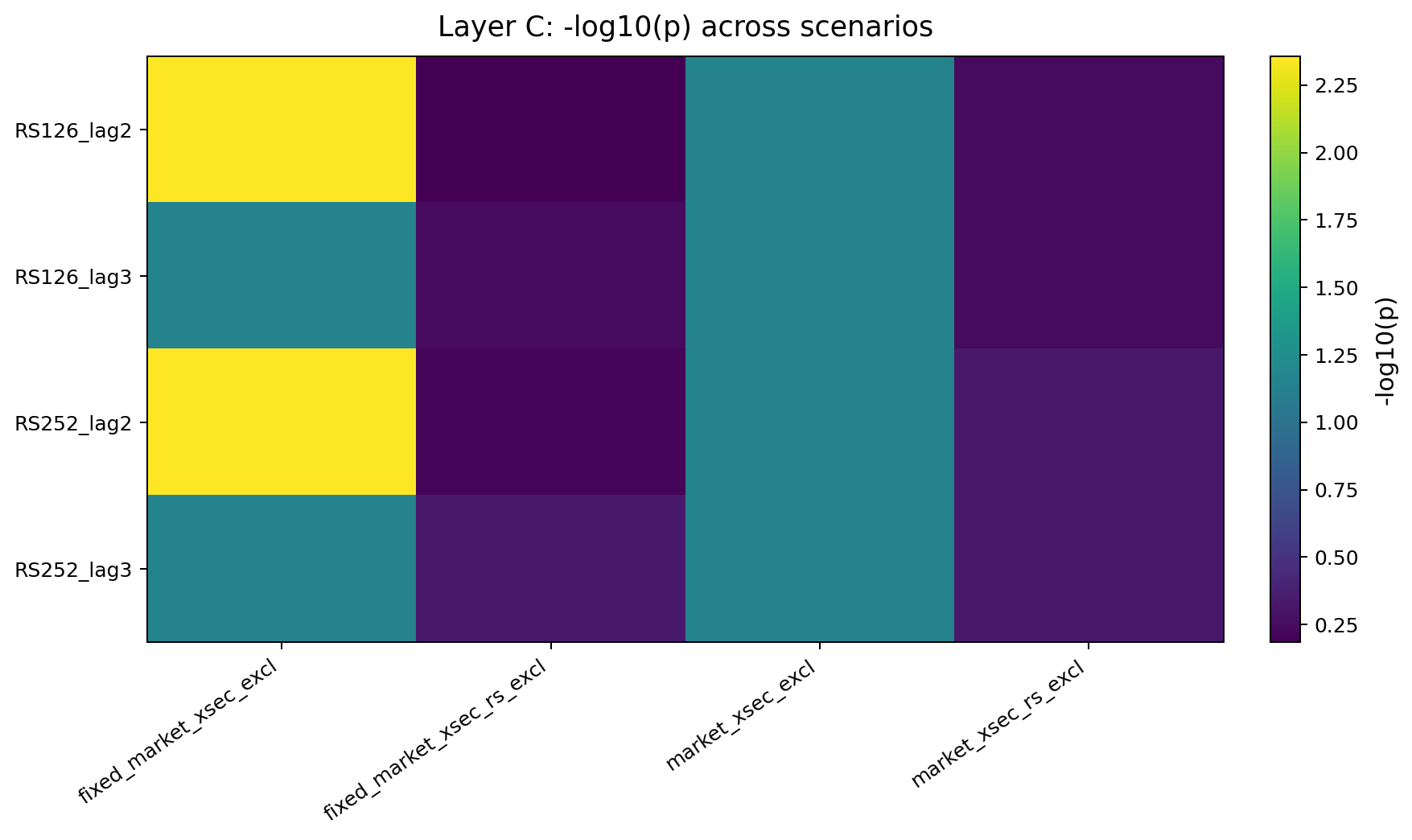}
  \caption{Layer~C robustness across RS windows and fixed lags ($-\log_{10}p$). File: \showtt{\robustdir/robust_summary_layerC_heatmap.png}.}
  \Description{Heatmap showing Layer C robustness across RS windows and fixed lags.}
  \label{fig:robustC}
\end{figure}

\noindent\textbf{VAR interpretation.}
\begin{figure}[H]
  \centering
  \safeinclude[width=\columnwidth]{var_main_fevd.png}
  \caption{VAR (main) FEVD. File: \showtt{\rundir/var_main_fevd.png}.}
  \Description{A plot showing the forecast error variance decomposition (FEVD) for the main VAR model.}
  \label{fig:varFEVD}
\end{figure}

\clearpage
\noindent\textbf{ML evaluation (classification).}
\begin{figure}[H]
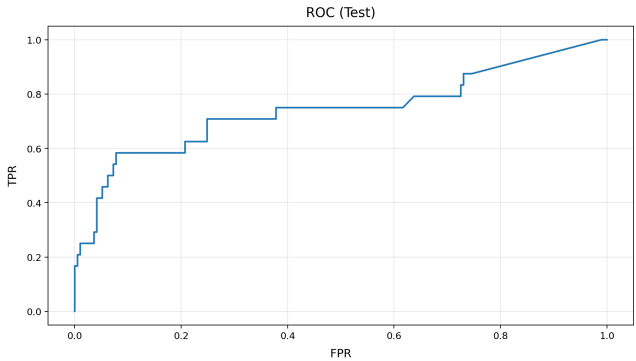

  \centering
  \safeinclude[width=\columnwidth]{cls_test_roc_curve.png}
  \caption{Test ROC curve (fixed threshold chosen on validation). File: \showtt{\rundir/cls_test_roc_curve.png}.}
  \Description{Test ROC curve showing precision-recall performance for the test set.}
  \label{fig:pr}
\end{figure}

\begin{figure}[H]
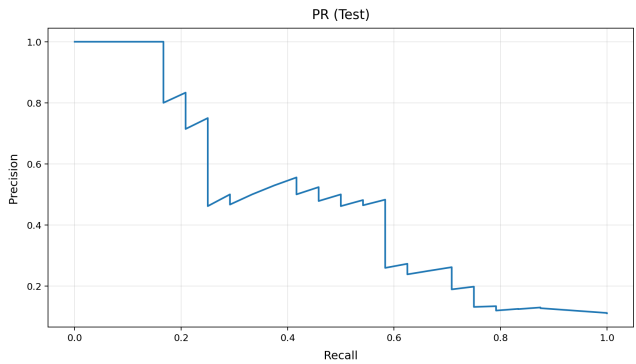

  \centering
  \safeinclude[width=\columnwidth]{cls_test_pr_curve.png}
  \caption{Test PR curve (fixed threshold chosen on validation). File: \showtt{\rundir/cls_test_pr_curve.png}.}
  \Description{Test PR curve showing precision-recall performance for the test set.}
  \label{fig:pr}
\end{figure}

\noindent\textit{Baseline comparison.}
Under class imbalance (positives $\approx$ top 15\%), PR–AUC is the primary discriminator.
While a calibrated Logit can post a high ROC–AUC (near 0.90), our XGBoost achieves a higher PR–AUC (0.47 vs.\ $\sim$0.44), reflecting earlier concentration of true positives at actionable recall.
We therefore emphasize PR–AUC and F1 at the validation–chosen, test–frozen threshold; full per–window metrics are in the artifacts.

\newpage
\noindent\textbf{Compact Layer~C table.}
\begin{table}[H]
\centering
\caption{Layer C robustness (p-values; bold = significant at $q{<}0.05$).}
\label{tab:robust-layerC}
\begin{tabular}{@{}c c c c c@{}}
\toprule
RS & Lag & Target & $p$-value & Sig. \\
\midrule
126 & 2 & Fixed/raw & \textbf{0.004} & \cmark \\
126 & 2 & Fixed/RS  & 0.655 &  \\
126 & 3 & Fixed/raw & 0.070 &  \\
126 & 3 & Fixed/RS  & 0.567 &  \\
252 & 2 & Fixed/raw & \textbf{0.004} & \cmark \\
252 & 2 & Fixed/RS  & 0.612 &  \\
252 & 3 & Fixed/raw & 0.070 &  \\
252 & 3 & Fixed/RS  & 0.462 &  \\
\bottomrule
\end{tabular}
\end{table}

\section{Discussion and Limitations}
Layer~A shows LV/volatility (\textsc{BNB}, \textsc{XEM}, \textsc{YFI}) Granger-causing returns, while returns drive residual vol/turnover. Layer~B links a non-market returns PC to vol-PCs. Layer~C shows vol-PCs leading our cross-sectional crowding target at short lags.  
Limitations: modest sample length; sensitivity to RS-window in Layer~C; potential structural breaks \cite{adelopo2025interconnectedness,hasan2022liquidity}. We therefore provide robustness grids, alternative targets (RS vs. raw; leave-$k$-out), baselines and DM/CI comparisons in the artifacts.

\noindent\textbf{Reproducibility.} All results are fully reproducible from the artifact folder \showtt{\rundir}: we fix chronological splits and random seeds, export \texttt{config.json} and \texttt{pipeline.log}, and release the aligned CSVs and figure scripts that regenerate every table and plot.

\section{Ethics}
Risk forecasts can influence capital allocation; misuse may amplify herding. We export thresholds/diagnostics/SHAP for transparency and position this as an early-warning tool, not a trading recommendation.

\clearpage
\bibliographystyle{ACM-Reference-Format}
\bibliography{reference}

\end{document}